\documentclass[conference]{IEEEtran}
\usepackage{amsmath,amssymb,amsfonts}
\usepackage{algorithmic}
\usepackage{algorithm}
\usepackage{graphicx}
\usepackage{textcomp}
\usepackage{xcolor}
\usepackage{url}
\usepackage{hyperref}
\usepackage{booktabs}
\usepackage{multirow}

\def\BibTeX{{\rm B\kern-.05em{\sc i\kern-.025em b}\kern-.08em
    T\kern-.1667em\lower.7ex\hbox{E}\kern-.125emX}}

\begin{document}

\title{Byzantine-Resilient Federated Learning via QUBO-Based Client Selection on Quantum Annealers}

\author{
\IEEEauthorblockN{Andras Ferenczi\IEEEauthorrefmark{1},
Sutapa Samanta\IEEEauthorrefmark{1},
Dagen Wang\IEEEauthorrefmark{1},
Jason Qizhe Qin\IEEEauthorrefmark{2}}
\IEEEauthorblockA{\IEEEauthorrefmark{1}American Express Co.}
\IEEEauthorblockA{\IEEEauthorrefmark{2}Columbia University}
}\maketitle

\begin{abstract}
Federated Learning (FL) is a well-established method for training a global model in a decentralized fashion, thus preserving the privacy of local data and eliminating the need for large centralized infrastructure. FL at scale runs the risk of malicious updates that can impact the global model. To mitigate such risks, a number of Byzantine-resilient aggregation methods have been introduced, one of the most popular ones being MultiKrum. Whereas this method scores the individual gradients against their nearest neighbors, it can easily miss malicious updates that preserve the statistical properties of honest updates. To mitigate such concerns, we propose a quantum annealing approach that reformulates client selection as a Quadratic Unconstrained Binary Optimization (QUBO) problem, encoding pairwise distances into a cost function solved by quantum annealers (QA). Unlike MultiKrum's greedy per-client scoring, the QUBO formulation jointly optimizes over all possible client subsets to find the mutually closest group of $m$ clients.
Our experiments at small scale consisting of 15 clients provided promising results for the most challenging Byzantine attacks. For example, the Advanced LIE attack was detected with 95.11\% accuracy with QUBO versus 81.33\% for classical MultiKrum on an MNIST-trained model and 97.78\% vs.\ 75.56\% on CIFAR-10. Whereas our QUBO method performs well against sophisticated attacks, it fares poorly against simpler attacks, which are handled well by classical MultiKrum. Our experiments show that the two methods complement each other well. Additionally, the QUBO method's quality degrades as we increase the number of clients.
To address these shortcomings, we introduce a MultiSignal ensemble that uses a dual-feature routing gate based on Euclidean and cosine Krum score gaps to classify attacks into four regimes and routes evasion attacks to a suspicion-penalized QUBO with agreement voting. The MultiSignal ensemble performs better than classical MultiKrum in all our experiments, with 95.3\% average detection accuracy versus 91.8\% for classical. The largest gains are achieved on Sparse Lie, where detection accuracy improves from 72.0\% to 95.2\%, a 23.2 percentage-point gain, and on Advanced Lie, where it increases from 80.4\% to 85.2\%, a 4.8 percentage-point gain. These results demonstrate that QUBO-based quantum annealing with the MultiSignal ensemble is a principled and scalable defense against the most challenging Byzantine strategies in federated learning.
\end{abstract}


\section{Introduction}

Federated Learning (FL)~\cite{McMahan2017, Kairouz2019, Yang2019} enables training of a global model by having clients (nodes) train local copies of the model and forwarding the gradient updates to a central server, which in turn, aggregates the updates and distributes the global model to clients for the next iteration of the training. This approach promotes privacy for the local data and eliminates the need for larger central infrastructure. When the clients are untrusted, nevertheless, the risk of having Byzantine updates can compromise the global model convergence~\cite{Blanchard2017}.

The existing state of the art, the Krum algorithm~\cite{Blanchard2017}, relies on calculation of pairwise Euclidean distances between gradients and selection of a single gradient that is closest to the majority. MultiKrum~\cite{Blanchard2017} extends this approach to select multiple clients. Both methods fail to detect more sophisticated attacks, such as the ``A Little Is Enough'' (Lie) attack family~\cite{baruch2019littleenoughcircumventingdefenses}, which maliciously alters the gradients while preserving the statistical properties of the updates.

To address these concerns, we propose formulating Byzantine client detection as a combinatorial optimization problem and solving it via quantum annealing. For this purpose, we encode the client selection objective as a Quadratic Unconstrained Binary Optimization (QUBO)~\cite{Kochenberger2014_QUBO} problem. This formulation enables \emph{global} optimization over all $\binom{n}{m}$ possible subsets, a search that is intractable for classical exhaustive methods but naturally suited to QAs~\cite{Albash2018}.

We make three contributions. First, we formulate the MultiKrum client selection problem as a QUBO expression, with a Lagrange penalty enforcing the selection of exactly~$m$ clients. We experiment at small scale with 15~clients across 13~distinct Byzantine attack types on MNIST~\cite{lecun-mnisthandwrittendigit-2010}, FashionMNIST~\cite{xiao2017fashionmnist}, and CIFAR-10~\cite{Krizhevsky09learningmultiple} data, and show that QUBO-based selection outperforms classical MultiKrum Byzantine update detection for the evasion category of attacks. Second, we introduce a threshold-based Cascaded Dual-QUBO method for larger instances with 100 clients where the performance of standard QUBO starts deteriorating. This method still fails to improve accuracy in detecting the Advanced Lie (ALIE) attack over classical MultiKrum. Third, we introduce a MultiSignal ensemble that computes Euclidean and cosine scores, constructs four regimes, and routes processing to either classical MultiKrum or a suspicion-penalized QUBO with agreement voting. At 100~clients on MNIST, the MultiSignal ensemble achieves 95.3\% average detection accuracy versus 91.8\% for classical MultiKrum, with its largest gains on Sparse Lie with a 23.2 percentage-point gain and ALIE with a 4.8 percentage-point gain, while preserving perfect detection on all outlier-producing attacks, including Clustered. We also provide a complexity analysis for classical MultiKrum, standard QUBO, cascade, and the MultiSignal approach.

This work complements our prior study on quantum embedding-based Byzantine detection~\cite{Ferenczi2025QE}, which uses parameterized quantum circuits for gradient feature extraction. While the quantum embedding approach leverages Hilbert space representations, the present work operates within the optimization paradigm of quantum annealing, offering a fundamentally different and complementary mechanism for robust aggregation.

\section{Related Work}

Krum~\cite{Blanchard2017}  selected the gradient that had its Euclidean distance closest to the rest. MultiKrum~\cite{ElMhamdi2018} extended the selection to $m$ clients. The paper~\cite{Yin2018} used coordinate-wise median and trimmed-mean estimators. DRACO~\cite{Chen2018} used redundant gradient computation for resilience. FLAME~\cite{cosine-flame} uses cosine-based clustering for backdoor defense, Sentinel~\cite{cosine-sentinel} combines cosine similarity filtering with bootstrap loss validation and normalization, and CosDefense~\cite{cosine-cosdefense} detects outliers via directional similarity. Dim-Krum~\cite{Zhang2022} applied Krum dimension-wise for NLP tasks. All of these classical solutions compute local or per-dimension optimization rather than global subset selection.

Wei et al.~\cite{Wei2024HQCBD} used Bender's decomposition framework to optimize FL scheduling in a distributed network. The same authors also applied quantum annealing to FL resource scheduling in~\cite{Wei2023QAFL}. Liu et al.~\cite{Liu2025QBAFL} proposed a quantum Byzantine agreement protocol for blockchain-based federated learning. Zhang et al.~\cite{Zhang2022QSA} used quantum key distribution to protect gradient updates. The paper~\cite{Xia2021QFLByzantine} tackled a scenario in which Byzantine participants send random or maliciously crafted unitaries. Subramanian and Chinnadurai~\cite{Subramanian2024} investigated the benefits of hybrid quantum-classical models when it comes to robustness in the face of adversarial updates.

In our prior work~\cite{Ferenczi2025QE}, we proposed using parameterized quantum circuits (Pauli feature maps) to embed projected gradients into a quantum Hilbert space, where inner products yield a quantum kernel matrix for Byzantine detection. We demonstrated improved detection of Byzantine updates by performing dimensionality reduction followed by projection into Hilbert space by means of feature maps.

\section{Background}

\subsection{Federated Learning}
In FL~\cite{McMahan2017}, $n$ clients collaboratively train a global model parameterized by $W$ under the coordination of a central server. In each round~$t$, the server broadcasts the current model~$W_t$ to all clients. Each client~$i$ trains locally on its private data and returns a gradient update $G_i = W_i' - W_t$. The server aggregates these updates and produces $W_{t+1}$. Assuming there are $f$ Byzantine clients out of the $n$ received by the central server, the goal is to generate a global model that is close to a 100\% honest aggregation.

\subsection{Krum and MultiKrum}
Let $G_1, G_2, \ldots, G_n \in \mathbb{R}^d$ be the gradient vectors. The distance between two gradients is
\begin{equation}
D(G_i, G_j) = \|G_i - G_j\|_2.
\end{equation}
For each update, Krum is defined as the sum of distances to its $n - f - 2$ nearest neighbors, i.e.,
\begin{equation}
\text{KRUM}(G_i) = \sum_{j \in \mathcal{N}_i} \|G_i - G_j\|_2^2,
\end{equation}
where $\mathcal{N}_i$ is the index set of the $n - f - 2$ closest gradients to $G_i$. Krum selects 
$$G_{\text{krum}} = \arg\min_i \text{KRUM}(G_i)\ .$$
MultiKrum method generalizes Krum by selecting $m$ gradients with the lowest Krum scores and averaging them as follows:
\begin{equation}
G_{\text{multi}} = \frac{1}{m} \sum_{i \in \mathcal{M}} G_i, \quad \mathcal{M} = \text{Top-}m\{\text{KRUM}(G_i)\}.
\end{equation}
Notably, MultiKrum does not optimize over subsets jointly. Any single Byzantine gradient that is close to its neighbors will evade detection.

\subsection{Quantum Annealing and QUBO}
Quantum annealers (QA)~\cite{Albash2018} are well-suited for optimization problems expressible as QUBO~\cite{Kochenberger2014_QUBO}:
\begin{equation}
\min_{x \in \{0,1\}^n} \sum_{i} Q_{ii}\, x_i + \sum_{i < j} Q_{ij}\, x_i x_j,
\label{eq:qubo_general}
\end{equation}
where $Q$ encodes both the objective and constraints. Solving an optimization problem encoded as QUBO form of Eq.~\eqref{eq:qubo_general} is equivalent to finding the ground state of an Ising Hamiltonian~\cite{Lucas2014}.

\section{Methodology}
We aim to replace the greedy MultiKrum algorithm with a method that holistically selects the most optimal subset based on the QUBO objectives and constraints.

\subsection{Step 1: Gradient Dimension Reduction}
The first step is to project a very large gradient vectors $G_i \in \mathbb{R}^d$ into a lower-dimensional space with dimension $k$. We use Importance-Weighted approach for this purpose. For each coordinate~$j$, we compute the variance $v_j = \text{Var}(\{G_{1,j}, \ldots, G_{n,j}\})$ across clients. We then select the top-$k$ coordinates by variance. This preserves dimensions where clients disagree most.
Let $\mathbf{p}_i \in \mathbb{R}^k$ denotes the selected gradient for client~$i$ after applying such dimension reduction.

\subsection{Step 2: Distance Matrix Computation}
Next, we compute a pairwise distance matrix $D \in \mathbb{R}^{n \times n}$ on the gradients with reduced dimension. We use two type of distance computation, cosine and Euclidean. The cosine distance is computed as follows:
\begin{equation}
D_{ij}^{\text{cosine}} = 1 - \frac{\mathbf{p}_i \cdot \mathbf{p}_j}{\|\mathbf{p}_i\| \, \|\mathbf{p}_j\| + \epsilon},
\label{eq:cosine_dist}
\end{equation}
Here $\epsilon = 10^{-8}$ prevents division by zero. Cosine distance measures directional dissimilarity and is invariant to gradient magnitude, making it well-suited for detecting attacks that preserve gradient norms but alter directions. The matrix is symmetric with $D_{ii}^{\text{cosine}} = 0$.
One the other hand, the Euclidean distance captures the magnitude dissimilarity which is calculated as follows:                                                
\begin{equation}                                                              
D_{ij}^{\text{euclid}} = \|\mathbf{p}_i - \mathbf{p}_j\|_2.
\label{eq:euclid_dist}
\end{equation}
For the dual-distance formulation (Section~\ref{sec:cascadeddualQUBO}), the
Euclidean matrix is normalized to $[0, 1]$:
\begin{equation}
\hat{D}_{ij}^{\text{euclid}} = \frac{D_{ij}^{\text{euclid}}}{\max_{k < l}
D_{kl}^{\text{euclid}}}.
\label{eq:euclid_norm}
\end{equation}

\subsection{Step 3: QUBO Formulation}
Let $x_i \in \{0,1\}$ indicates whether client~$i$ is selected. We seek to select $m = n - f$ clients whose projected gradients are mutually closest. Our objective is to minimize the total pairwise distance among selected clients, i.e.,
\begin{equation}
\min_{x \in \{0,1\}^n} \sum_{i=1}^{n} \left(\sum_{j=1}^{n} D_{ij}\right) x_i - 2 \sum_{i < j} D_{ij}\, x_i x_j.
\label{eq:qubo_obj}
\end{equation}
To enforce exactly $m$ selections, we add a Lagrange penalty of the form:
\begin{equation}
\lambda \left(\sum_{i=1}^{n} x_i - m\right)^2,
\label{eq:constraint}
\end{equation}
where $\lambda$ is the Lagrange multiplier. We set the value $\lambda = 10 \times \max_{i,j} D_{ij}$, ensuring the constraint penalty dominates the objective when violated. 
Expanding Eq.~\eqref{eq:qubo_obj}, and Eq.~\eqref{eq:constraint} and using the property of binary variables, $x_i^2 = x_i$, and $m^2$ is a constant term, irrelevant to the optimization, the elements of QUBO can be written as follows:
\begin{align}
Q_{ii} &= \sum_{j=1}^{n} D_{ij} + \lambda(1 - 2m), \label{eq:final_diag}\\
Q_{ij} &= -2D_{ij} + 2\lambda, \quad \text{for } i < j. \label{eq:final_offdiag}
\end{align}

We solve the QUBO using a simulated annealing sampler with 1000 reads and extract client indices~$S$ for small instances. If $|S| \neq m$ we perform post-processing to add or remove clients based on their distance sums to the current selection.
The global model is updated as:
\begin{equation}
W_{t+1} = W_t + \eta \cdot \frac{1}{|S|} \sum_{i \in S} G_i.
\end{equation}

\subsection{Cascaded Dual-QUBO}
\label{sec:cascadeddualQUBO}
For large client populations ($n \geq 100$), the solution above faces scaling issues and there is a higher likelihood that Byzantine updates cluster tightly thus confusing our optimization algorithm. We attempt to resolve this using a Cascaded Dual-QUBO approach. We define Krum score gap $\Delta$ as follows:
\begin{equation}
\Delta = \frac{\min_{i \in \mathcal{R}} s_i - \max_{j \in \mathcal{S}} s_j}{\text{std}(\{s_1, \ldots, s_n\})},
\label{eq:routing_gap}
\end{equation}
where $\mathcal{S}$ and $\mathcal{R}$ are the selected and rejected client sets under that metric's Krum scoring, and $s_i$ is the Krum score for client~$i$. We introduce a gap threshold $\tau$. When Euclidean Krum score gap is above this threshold, classical MultiKrum is good enough. When the gap is below the threshold, classical method fails. In that case we build a QUBO using equal blend of normalized Euclidean and cosine distances defined as follows:
\begin{equation}
D_{ij}^{\text{dual}} = \alpha \hat{D}_{ij}^{\text{euclid}} + (1 - \alpha)  D_{ij}^{\text{cosine}},
\label{eq:dual_dist}
\end{equation}
where $\hat{D}^{\text{euclid}}$ is the Euclidean distance matrix normalized to $[0, 1]$ and $\alpha = 0.5$. Then solve it using BQM hybrid quantum annealing solver.

\subsection{MultiSignal Ensemble}
\label{sec:multisignal}
Our initial Cascaded Dual-QUBO approach for 100 client instances, however, did not work for ALIE.
We address this concern with MultiSignal ensemble that comprises of a dual-feature routing gate, a suspicion-penalized QUBO, and and an agreement voting.

\subsubsection*{Stage 1: Dual-Feature Routing Gate} We compute the normalized Euclidean Krum score gap $\Delta_E$ on full gradients and the normalized cosine Krum score gap $\Delta_C$ on importance-weighted projected gradients using Eq.~\eqref{eq:routing_gap}. 
We also define cutoffs $\tau_c$ and $\tau_E$ for cosine and Euclidean gap respectively.
Based on these two gap values we categorize the outcomes as follows: (1) Outlier when $\Delta_E > \tau_E$ and $\Delta_C > \tau_E$, (2) Clustered when $\Delta_E \leq \tau_E$ and $\Delta_C > \tau_C$, (3) Magnitude when $\Delta_E > \tau_E$ and $\Delta_C \leq \tau_E$, and (4) Evasion when $\Delta_E \leq \tau_E$ and $\Delta_C \leq \tau_E$. For the first three categories, classical multi-Kurm works well. For the fourth category, both the gaps are low and a suspicion-penalized
QUBO with agreement voting is more beneficial. 
Based on extensive experimentation we settled with $\tau_E = 0.2$ and $\tau_C = 0.5$. In our experiments we were unable to produce Outlier and Clustered scenarios. These are hypothetical scenarios for future-proofing our system.

\subsubsection*{Stage 2: Suspicion-Penalized QUBO} 
While the standard QUBO objective in Eq.~\eqref{eq:qubo_obj} encourages client mutual proximity, it cannot detect ALIE in which gradients are extremely close to the honest mean. Suspicion-Penalized QUBO discourages selecting pairs of clients whose gradients are suspiciously similar. First, we construct a dual distance matrix on importance-weighted projected gradients following Eq.~\eqref{eq:dual_dist}.
The suspicion threshold $\tau_s$ is $p$-th percentile of the upper-triangular entries of $D^{\text{dual}}$, with $p = 10$. The modified quadratic coefficients become:
\begin{equation}
Q_{ij}^{\text{susp}} = -2D_{ij}^{\text{dual}} + w_s \cdot \max(0,\, \tau_s - D_{ij}^{\text{dual}}),
\label{eq:suspicion_penalty}
\end{equation}
where $w_s = 10.0$ is the suspicion weight. For pairs with $D_{ij}^{\text{dual}} < \tau_s$, the penalty term overpowers the incentive.

\subsubsection*{Stage 3: Agreement Voting} 
In the evasion regime we run both classical MultiKrum and QUBO in parallel. As both Euclidean and cosine gaps are low, classical is given a low confidence. We produce four categories: \emph{agreed-accept} when both methods select, \emph{agreed-reject} when both methods reject, \emph{QUBO-only} when QUBO selects but classical rejects, and \emph{classical-only} when classical selects but QUBO rejects. A blended rank score based on normalized cosine Krum scores, with bonuses for agreed-accept and penalties for agreed-reject and classical-only, determines the final selection of $m$~clients.
Algorithm~\ref{alg:qubo_agg} summarizes the complete procedure.

\begin{algorithm}[t]
\caption{MultiSignal Ensemble for Byzantine-Robust Aggregation}
\label{alg:qubo_agg}
\begin{algorithmic}[1]
\REQUIRE Clients to select $m$, thresholds $\tau_E, \tau_C$, projection dim $k$
\FOR{each FL round $t = 1, 2, \ldots$}
    \STATE Broadcast $W_t$; receive gradients $G_1, \ldots, G_n$
    \STATE Project: $\mathbf{p}_i \leftarrow \text{Project}(G_i, k)$ for all $i$
    \STATE Compute Euclidean gap $\Delta_E$ and cosine gap $\Delta_C$ (Eq.~\eqref{eq:routing_gap})
    \IF{$\Delta_E > \tau_E$ and $\Delta_C > \tau_E$}
        \STATE \textbf{OUTLIER}: $S \leftarrow \text{ClassicalMultiKrum}(G, m)$
    \ELSIF{$\Delta_E \leq \tau_E$ and $\Delta_C > \tau_C$}
        \STATE \textbf{CLUSTERED}: $S \leftarrow \text{ClassicalMultiKrum}(G, m)$
    \ELSIF{$\Delta_E > \tau_E$ and $\Delta_C \leq \tau_E$}
        \STATE \textbf{MAGNITUDE}: $S \leftarrow \text{ClassicalMultiKrum}(G, m)$
    \ELSE
        \STATE \textbf{EVASION}: $S_c \leftarrow \text{ClassicalMultiKrum}(G, m)$
        \STATE Construct suspicion QUBO (Eqs.~\eqref{eq:dual_dist}--\eqref{eq:suspicion_penalty})
        \STATE $S_q \leftarrow \text{AnnealSolve}(Q^{\text{susp}})$
        \STATE $S \leftarrow \text{AgreementVote}(S_c, S_q, m)$
    \ENDIF
    \STATE $W_{t+1} \leftarrow W_t + \eta \cdot \frac{1}{|S|} \sum_{i \in S} G_i$
\ENDFOR
\end{algorithmic}
\end{algorithm}

\section{Byzantine Attack Models}
We evaluate against 13 distinct Byzantine attack strategies listed in Table~\ref{tab:attack_models}. Let $\mu$ and $\sigma$ denote the coordinate-wise mean and standard deviation of honest client gradients. Let $G_i$ denote an honest gradient and $\widetilde{G}$ denotes the Byzantine update crafted from $(\mu,\sigma)$.
\begin{table*}[ht]
\caption{Byzantine attack models used in evaluation.}
\label{tab:attack_models}
\centering
\small
\begin{tabular}{p{2.8cm} p{7.0cm} p{5.5cm}}
\toprule
Attack & Byzantine update $\widetilde{G}$ & Notes / parameters \\
\midrule
Gaussian Noise &
$\widetilde{G} \sim \mathcal{N}(0,\alpha^2 I)$
& scaled random noise \\

Sign flip &
$\widetilde{G} = -\mu$
& negate the honest mean \\

Scaling &
$\widetilde{G} = \alpha \mu + \mathcal{N}(0,0.1^2 I)$
& add noise to scaled mean \\

Targeted &
$\widetilde{G}_j = \begin{cases} \mu_j + 5\sigma_j & \text{w.p.\ } 0.1, \\ \mu_j & \text{otherwise.} \end{cases}$
& Modify a random 10\% of coordinates \\

Clustered &
$\widetilde{G} = \mu + 2\beta\sigma$
& $\beta$ Byzantine clients \\

Same Value &
$\widetilde{G} = \mu + 3\sigma$
& identical shifted updates\\

Lie Attack &
$\widetilde{G} = \mu - z_{\max}\sigma$
& $z_{\max}\in\{0.5,1.0,1.5,2.0\}$ \\

Blatant Lie &
$\widetilde{G} = \mu + d\,z\,\sigma + \mathcal{N}(0,\eta^2 I)$
& $d\in\{-1,1\}, z\sim \text{Uniform}(1.75,3.25)$, $\eta\sim \text{Uniform}(0.1,0.4)\times\mathrm{mean}(\sigma)$ \\

Advanced Lie (ALIE) &
$\widetilde{G}_j = \begin{cases} \mu_j \pm z_{\text{large}} \cdot \sigma_j & \text{w.p.\ } 0.2, \\ \mu_j & \text{otherwise,} \end{cases}$
& $z_{\mathrm{large}}\sim \text{Uniform}(3,8)$; 70\% negative \\

Sparse Lie &
$\widetilde{G}_j = \begin{cases} \mu_j - z_{\text{ext}} \cdot \sigma_j & j \in \text{Top-5\%}, \\ \mu_j & \text{otherwise,} \end{cases}$
& $z_{\mathrm{ext}}\sim \text{Uniform}(5,15)$ \\

Label Flip & $y'=(y+1)\bmod C$ &
train on shifted data labels
 \\

Shuffle &
$\widetilde{G} = \mathrm{Shuffle}(G_{\mathrm{honest}})$
& permute coordinates randomly\\

Stealthy &
$\widetilde{G} = \mu + \mathcal{N}(0,0.05^2 I)$
& imperceptible noise \\
\bottomrule
\end{tabular}
\end{table*}

These attacks range from easily detectable outlier-producing strategies like Gaussian Noise and Scale to sophisticated methods specifically designed to evade distance-based defenses like Lie family and data poisoning Label Flip, providing a comprehensive evaluation of robustness.

\section{Experimental Setup}

\subsection{Datasets and Models}
We evaluate on three standard image classification benchmarks. MNIST is the handwritten digit dataset consisting of 28$\times$28 grayscale images categorized into 10 classes. This dataset is paired with a fully connected neural network  with $784\to200\to10$ architecture, ReLU activations, and approximately 159K parameters. Data is partitioned across clients using non-IID splits. FashionMNIST is the Zalando clothing article dataset~\cite{xiao2017fashionmnist} consists of 28$\times$28 grayscale images with 10 classes. This dataset serves as a more challenging drop-in replacement for MNIST. CIFAR-10 is  the natural image dataset  with 32$\times$32 RGB images consist of 10 classes. This dataset uses a compact CNN consisting of two convolutional layers 3$\to$4 and 4$\to$8 channels; 3$\times$3 kernels, and max-pooling, followed by two fully connected layers with 512$\to$32$\to$10 architecture, totaling approximately 17K parameters. The metrics used to evaluate performance of different methods are defined in Table~\ref{tab:metricsdef}.

\subsection{15-Client Configuration}
We use 12 honest and 3 Byzantine clients ($n = 15$, $f = 3$) and select $m = n-f = 12$ clients per round. We run 30 FL rounds. Local training uses SGD with learning rate 0.01 and weight decay $10^{-4}$, and we apply gradient clipping with max norm 1.0.
We use an importance-weighted projection of the gradients with dimension $k = 1000$ and compute distance matrix using cosine distance using Eq.~\eqref{eq:cosine_dist}. The QUBO is solved using the Neal simulated annealing sampler with 1000 reads. 
Classical MultiKrum uses the same importance-weighted projection and squared Euclidean distances on projected gradients, with $k = n - f - 2 = 10$ nearest neighbors for Krum scoring.

\subsection{100-Client QA Hybrid Configuration}
To evaluate scalability, we run a second set of experiments at a larger scale with 80 honest clients and  20 Byzantine clients, i.e., $n = 100$, $f = 20$, selecting $m = n - f = 80$ clients per round. We use Cascaded Dual-QUBO, and  MultiSignal ensemble method using the LeapHybridSampler which is hybrid quantum-classical quantum annealer. 
We perform 5 FL rounds using the MNIST dataset. We use a subset of 6 representative Byzantine attacks out of 13 to limit computational cost both on classical and quantum hardware. We use the same importance-weighted projection with $k = 1000$, and SGD with learning rate 0.01. The classical baseline uses the same 100-client configuration with standard MultiKrum selection of $k_{\text{nn}} = n - f - 2 = 78$ nearest neighbors.

\begin{table}[ht]
\caption{Definitions of metrics used to compare FL performance. TP, FN, TN are abbreviations for True Positive, False Negative, and True Negatives respectively. All metrics are averaged over the 30 FL rounds.
}
\label{tab:metricsdef}
\begin{tabular}{c|c}
    \hline
    Metrics & Definitions\\
    \hline
    Detection Accuracy &  $(TP + TN) / n$\\
    F1 Score & $2  TP / (2 TP + FP + FN)$\\
    Byzantine Rejection Rate & $TP / f$\\
    Honest Retention Rate & $TN / (n - f)$\\
    \hline
\end{tabular}
\end{table}

\section{Results and Analysis}
\subsection{15 clients FL results}
\label{sec:15client_results}
Table~\ref{tab:metric_summary} summarizes the average detection accuracy and F1 score for both aggregators across all 13 attack types for MNIST data. Classical MultiKrum excels on nine of 13 attacks such as Gaussian Noise, Scale, Targeted, ALIE, Clustered, Same Value, Blatant Lie, Label Flip, and Stealthy which are easily identifiable by per-client Krum scoring with squared Euclidean distances. In contrast, the more subtle Lie attack family, i.e., ALIE is better handled by the QUBO-based approach, achieving 95.11\% detection accuracy versus 81.33\% for classical MultiKrum, with an F1 score of 87.78\% versus 53.33\%. Standard Lie detection improves from 72.44\% to 86.67\% with F1 score increase from 31.11\% to 66.67\%. Sparse Lie accuracy improves from 66.67\% to 80.89\% with F1 score increasing from  16.67\% to 52.22\%. For Sign Flip, classical MultiKrum fails entirely with TP = 0 and detection accuracy = 60\%, while QUBO formulation using simulated annealing achieves TP $\approx$ 0.83 and a detection accuracy of 71.11\%. Taken together, these complementary strengths motivate an ensemble method as the most effective overall solution.
\begin{table}[t]
\caption{MNIST: Average Detection Accuracy and F1 Score across 30 rounds. Bold indicates the better method for each attack.}
\label{tab:metric_summary}
\centering
\small
\begin{tabular}{l|cc|cc}
\toprule
 & \multicolumn{2}{c|}{\textbf{Detection Acc.\ (\%)}} & \multicolumn{2}{c}{\textbf{F1 Score (\%)}} \\
\textbf{Attack} & Classical & QUBO & Classical & QUBO \\
\midrule
Gaussian Noise    & \textbf{100.00} & 60.00 & \textbf{100.00} & 0.00  \\
Sign Flip         & 60.00 & \textbf{71.11} & 0.00  & \textbf{27.78} \\
Scale             & \textbf{100.00} & 74.67 & \textbf{100.00} & 36.67 \\
Targeted          & \textbf{100.00} & 86.67 & \textbf{100.00} & 66.67 \\
Clustered         & \textbf{100.00} & 85.78 & \textbf{100.00} & 64.44 \\
Same Value        & \textbf{100.00} & 85.78 & \textbf{100.00} & 64.44 \\
Lie               & 72.44 & \textbf{86.67} & 31.11 & \textbf{66.67} \\
Blatant Lie       & \textbf{100.00} & 83.11 & \textbf{100.00} & 57.78 \\
ALIE      & 81.33 & \textbf{95.11} & 53.33 & \textbf{87.78} \\
Sparse Lie        & 66.67 & \textbf{80.89} & 16.67 & \textbf{52.22} \\
Label Flip        & \textbf{100.00} & 60.00 & \textbf{100.00} & 0.00  \\
Shuffle           & \textbf{69.33} & 60.00 & \textbf{23.33} & 0.00  \\
Stealthy          & \textbf{100.00} & 82.67 & \textbf{100.00} & 56.67 \\
\bottomrule
\end{tabular}
\end{table}
Table~\ref{tab:cifar_summary} presents the corresponding results on CIFAR-10, confirming the MNIST findings hold for more complex datasets.
\begin{table}[t]
\caption{CIFAR-10: Average Detection Accuracy and F1 Score across 30 rounds. Bold indicates the better method for each attack.}
\label{tab:cifar_summary}
\centering
\small
\begin{tabular}{l|cc|cc}
\toprule
 & \multicolumn{2}{c|}{\textbf{Detection Acc.\ (\%)}} & \multicolumn{2}{c}{\textbf{F1 Score (\%)}} \\
\textbf{Attack} & Classical & QUBO & Classical & QUBO \\
\midrule
Gaussian Noise    & \textbf{100.00} & 60.00 & \textbf{100.00} & 0.00  \\
Sign Flip         & 61.33 & \textbf{70.67} & 3.33  & \textbf{26.67} \\
Scale             & \textbf{100.00} & 87.11 & \textbf{100.00} & 67.78 \\
Targeted          & \textbf{100.00} & 96.44 & \textbf{100.00} & 91.11 \\
Clustered         & \textbf{100.00} & 75.11 & \textbf{100.00} & 37.78 \\
Same Value        & \textbf{100.00} & 86.67 & \textbf{100.00} & 66.67 \\
Lie               & \textbf{76.00} & 73.33 & \textbf{40.00} & 33.33 \\
Blatant Lie       & \textbf{100.00} & 69.33 & \textbf{100.00} & 23.33 \\
ALIE      & 75.56 & \textbf{97.78} & 38.89 & \textbf{94.44} \\
Sparse Lie        & 72.44 & \textbf{73.33} & 31.11 & \textbf{33.33} \\
Label Flip        & \textbf{100.00} & 60.00 & \textbf{100.00} & 0.00  \\
Shuffle           & 70.22 & \textbf{90.67} & 25.56 & \textbf{76.67} \\
Stealthy          & \textbf{100.00} & 96.00 & \textbf{100.00} & 90.00 \\
\bottomrule
\end{tabular}
\end{table}
Table~\ref{tab:fmnist_summary} presents FashionMNIST results. Using the same MLP architecture as MNIST but with more complex visual features, FashionMNIST closely mirrors the MNIST detection pattern, confirming that the complementary strengths are robust to gradient landscape complexity within the same architecture.
\begin{table}[t]
\caption{FashionMNIST: Average Detection Accuracy and F1 Score across 30 rounds. Bold indicates the better method for each attack.}
\label{tab:fmnist_summary}
\centering
\small
\begin{tabular}{l|cc|cc}
\toprule
 & \multicolumn{2}{c|}{\textbf{Detection Acc.\ (\%)}} & \multicolumn{2}{c}{\textbf{F1 Score (\%)}} \\
\textbf{Attack} & Classical & QUBO & Classical & QUBO \\
\midrule
Gaussian Noise    & \textbf{100.00} & 60.00 & \textbf{100.00} & 0.00  \\
Sign Flip         & 60.00 & \textbf{69.78} & 0.00  & \textbf{24.44} \\
Scale             & \textbf{100.00} & 60.00 & \textbf{100.00} & 0.00  \\
Targeted          & \textbf{100.00} & 74.22 & \textbf{100.00} & 35.56 \\
Clustered         & \textbf{100.00} & 73.33 & \textbf{100.00} & 33.33 \\
Same Value        & \textbf{100.00} & 73.33 & \textbf{100.00} & 33.33 \\
Lie               & 74.67 & \textbf{86.67} & 36.67 & \textbf{66.67} \\
Blatant Lie       & \textbf{100.00} & 88.00 & \textbf{100.00} & 70.00 \\
ALIE      & 81.33 & \textbf{95.11} & 53.33 & \textbf{87.78} \\
Sparse Lie        & 65.78 & \textbf{78.22} & 14.44 & \textbf{45.56} \\
Label Flip        & \textbf{100.00} & 60.00 & \textbf{100.00} & 0.00  \\
Shuffle           & \textbf{70.22} & 60.00 & \textbf{25.56} & 0.00  \\
Stealthy          & \textbf{100.00} & 86.22 & \textbf{100.00} & 65.56 \\
\bottomrule
\end{tabular}
\end{table}



Comparing Tables~\ref{tab:metric_summary},~\ref{tab:cifar_summary}, and~\ref{tab:fmnist_summary} reveals that the complementary pattern generalizes across all three datasets, with dataset- and architecture-specific effects. Table~\ref{tab:evasion_summary} isolates the five evasion attacks where classical MultiKrum falls below perfect detection, confirming that the QUBO formalism advantage on these attacks is consistent across datasets.
\begin{table*}[t]
\caption{Cross-dataset comparison on evasion attacks: Detection Accuracy (\%) for Classical MultiKrum vs.\ QUBO. These five attacks are the only ones where classical detection falls below 100\%. Bold indicates the better method; $\Delta$ shows the QUBO advantage in percentage points.}
\label{tab:evasion_summary}
\centering
\small
\begin{tabular}{l|ccc|ccc|ccc}
\toprule
 & \multicolumn{3}{c|}{\textbf{MNIST}} & \multicolumn{3}{c|}{\textbf{CIFAR-10}} & \multicolumn{3}{c}{\textbf{FashionMNIST}} \\
\textbf{Attack} & Classical & QUBO & $\Delta$ & Classical & QUBO & $\Delta$ & Classical & QUBO & $\Delta$ \\
\midrule
ALIE  & 81.33 & \textbf{95.11} & +13.8 & 75.56 & \textbf{97.78} & +22.2 & 81.33 & \textbf{95.11} & +13.8 \\
Lie           & 72.44 & \textbf{86.67} & +14.2 & \textbf{76.00} & 73.33 & $-$2.7 & 74.67 & \textbf{86.67} & +12.0 \\
Sparse Lie    & 66.67 & \textbf{80.89} & +14.2 & 72.44 & \textbf{73.33} & +0.9 & 65.78 & \textbf{78.22} & +12.4 \\
Sign Flip     & 60.00 & \textbf{71.11} & +11.1 & 61.33 & \textbf{70.67} & +9.3 & 60.00 & \textbf{69.78} & +9.8 \\
Shuffle       & \textbf{69.33} & 60.00 & $-$9.3 & 70.22 & \textbf{90.67} & +20.4 & \textbf{70.22} & 60.00 & $-$10.2 \\
\midrule
\textbf{Average (5 attacks)} & 69.95 & \textbf{78.76} & +8.8 & 71.11 & \textbf{81.16} & +10.0 & 70.40 & \textbf{77.96} & +7.6 \\
\bottomrule
\end{tabular}
\end{table*}

For ALIE, the QUBO advantage is consistent across the three datasets, achieving 95.11\% on both MNIST and FashionMNIST vs.\ 81.33\% using classical MultiKrum, with an even larger gap on CIFAR-10 where we see 97.78\% vs.\ 75.56\%, a positive 22.2 percent gap. The identical MNIST and FashionMNIST results, despite different data complexity, suggest that the advantage stems from the optimization mechanism rather than the data distribution, while the larger CIFAR-10 gap is consistent with CNN-based architectures being more sensitive to directional changes that cosine distances can capture in the QUBO formulation. For the Lie attack, architecture matters as we see QUBO formalism outperforms classical MultiKrum on the MLP-based datasets, i.e., MNIST and FashionMNIST, but not on CIFAR-10, which is consistent with lower parameter dimensionality (~17K for MNIST and Fashion MNIST vs.\ ~159K for CIFAR-10) producing larger per-coordinate shifts that are more detectable by classical MultiKrum's Euclidean scoring. For Shuffle, classical MultiKrum has a slight edge on MNIST and FashionMNIST, whereas on CIFAR-10 QUBO performs better as shuffling disrupts CNN filter-level coherence. Finally, FashionMNIST closely matches MNIST across all attacks, reinforcing that detection performance depends primarily on model architecture and attack type rather than visual dataset complexity.

Overall, the three-dataset comparison (Table~\ref{tab:evasion_summary}) confirms that QUBO-based optimization provides its greatest value against attacks that exploit the limitations of per-client distance scoring such as ALIE, Lie, Sparse Lie, and Sign Flip, with an average improvement of 8 to 10 percentage points on these five evasion attacks, consistent across architectures and amplified on structured CNN gradient spaces.

\begin{figure}[t]
\centerline{\includegraphics[width=\linewidth]{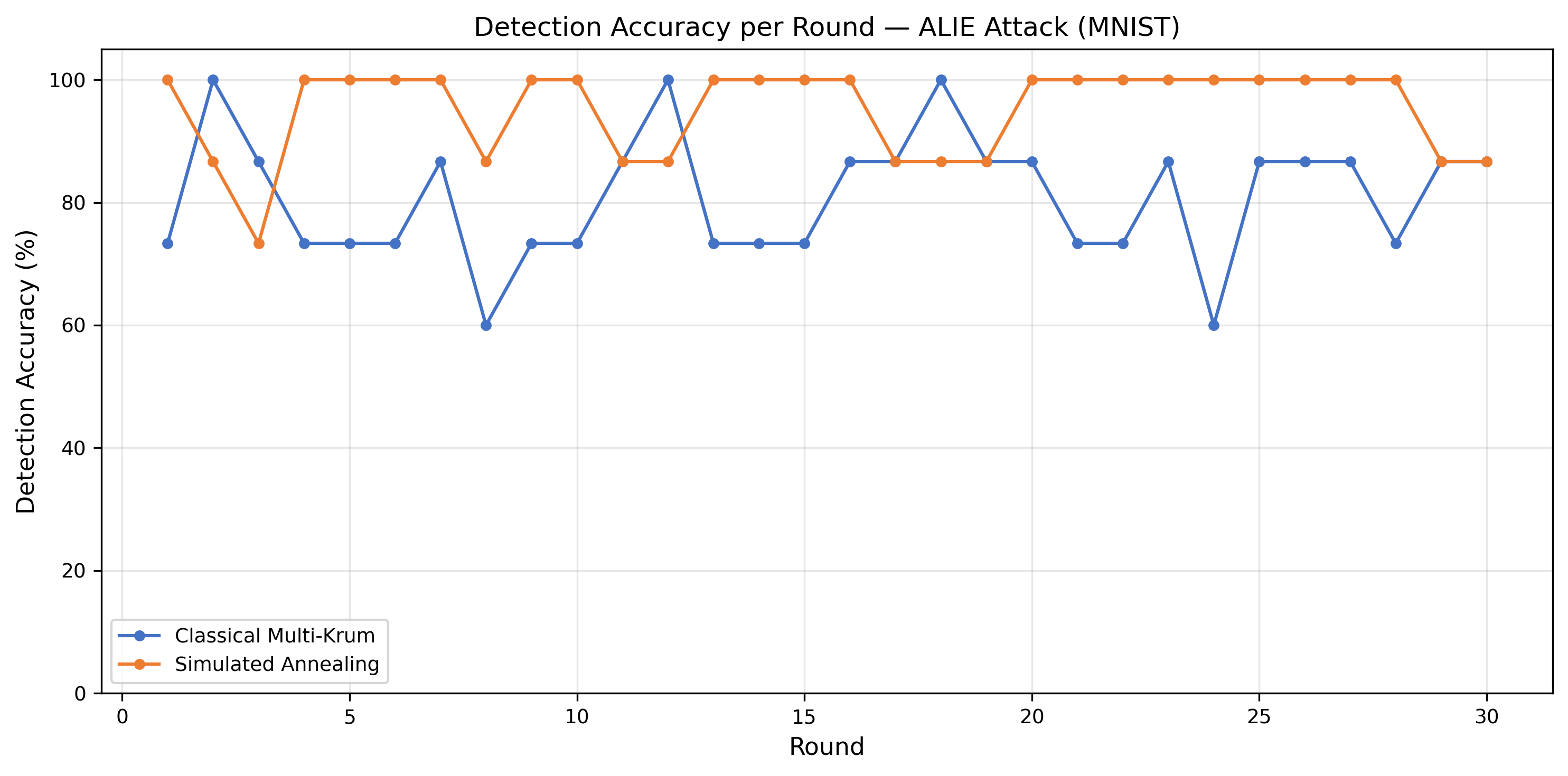}}
\caption{MNIST: Round-by-round detection under ALIE attack.}
\label{fig:advanced_lie}
\end{figure}

\begin{figure}[t]
\centerline{\includegraphics[width=\linewidth]{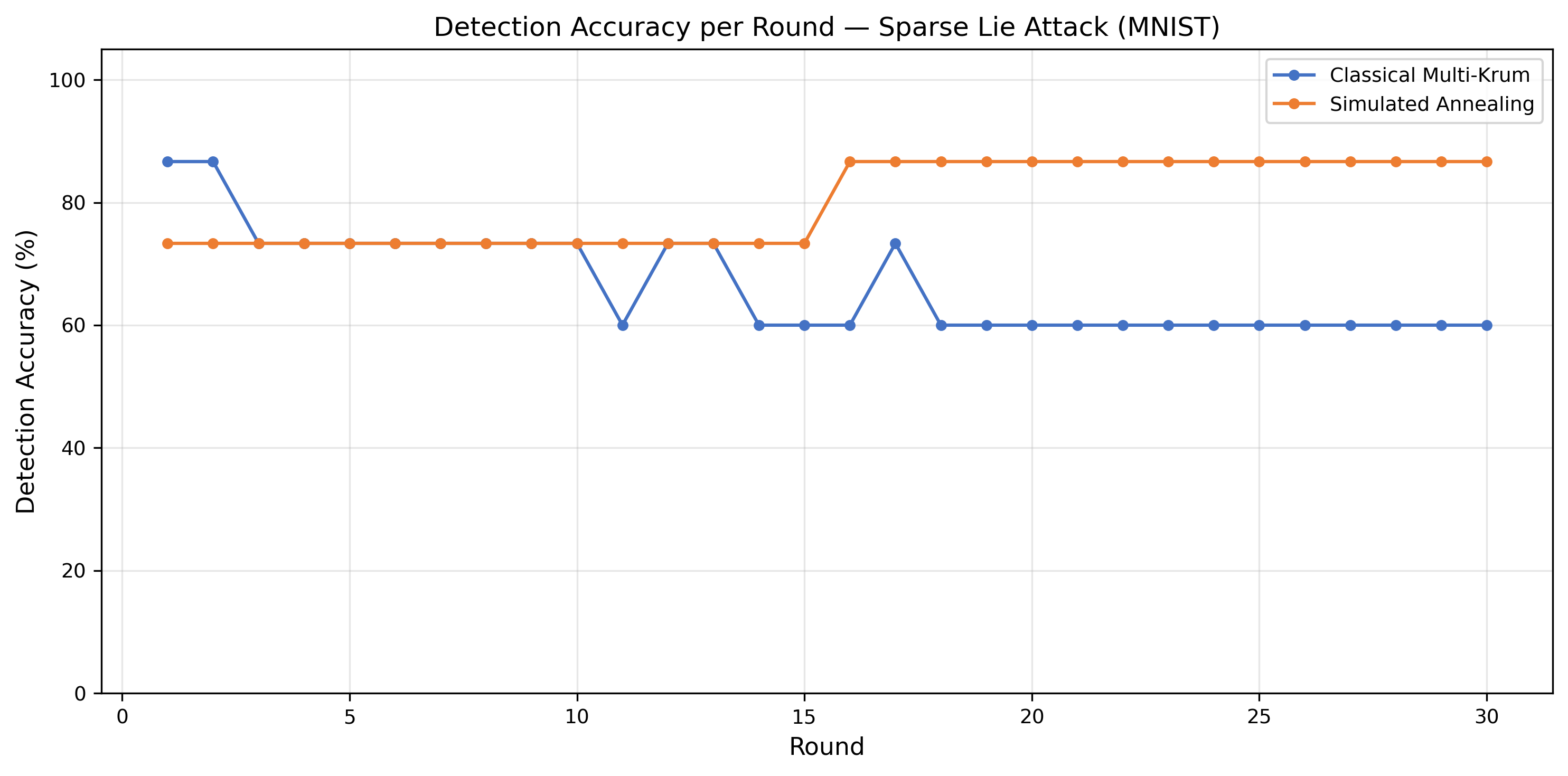}}
\caption{MNIST: Round-by-round detection under Sparse Lie attack.}
\label{fig:sparse_lie}
\end{figure}

\begin{figure}[t]
\centerline{\includegraphics[width=\linewidth]{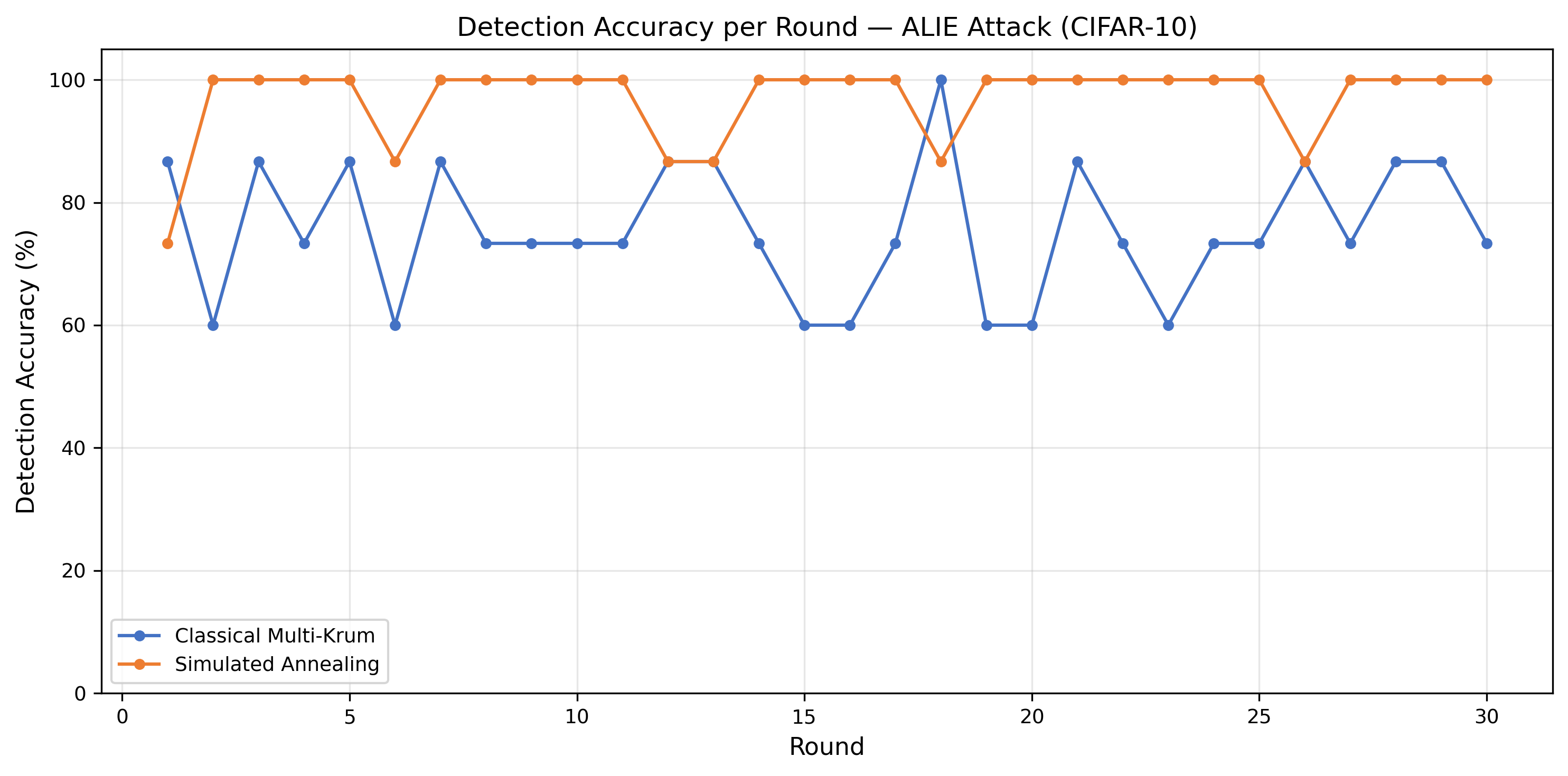}}
\caption{CIFAR-10: Round-by-round detection under ALIE attack.}
\label{fig:cifar_advanced_lie}
\end{figure}

\begin{figure}[t]
\centerline{\includegraphics[width=\linewidth]{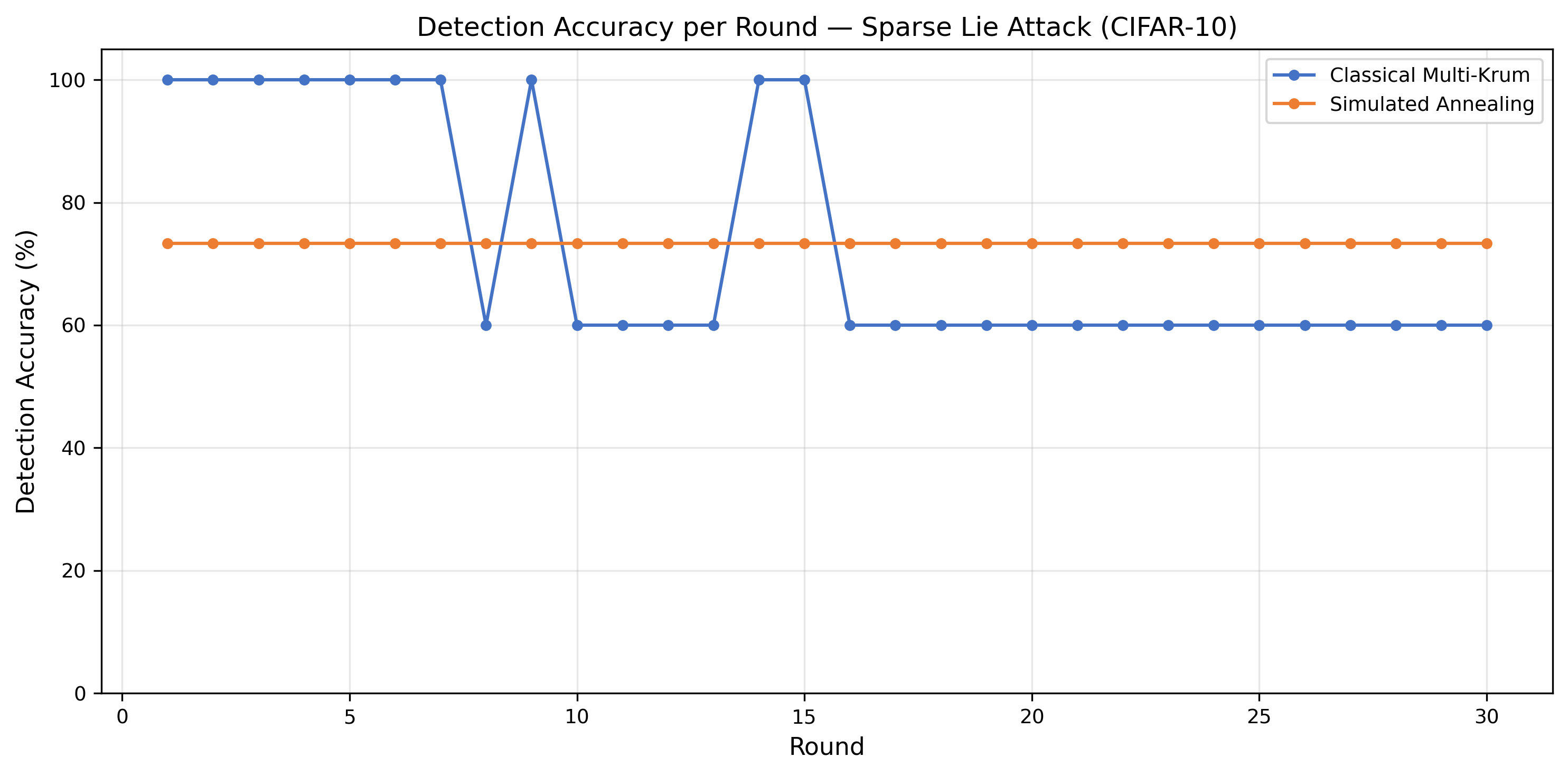}}
\caption{CIFAR-10: Round-by-round detection under Sparse Lie attack.}
\label{fig:cifar_sparse_lie}
\end{figure}

\begin{figure}[t]
\centerline{\includegraphics[width=\linewidth]{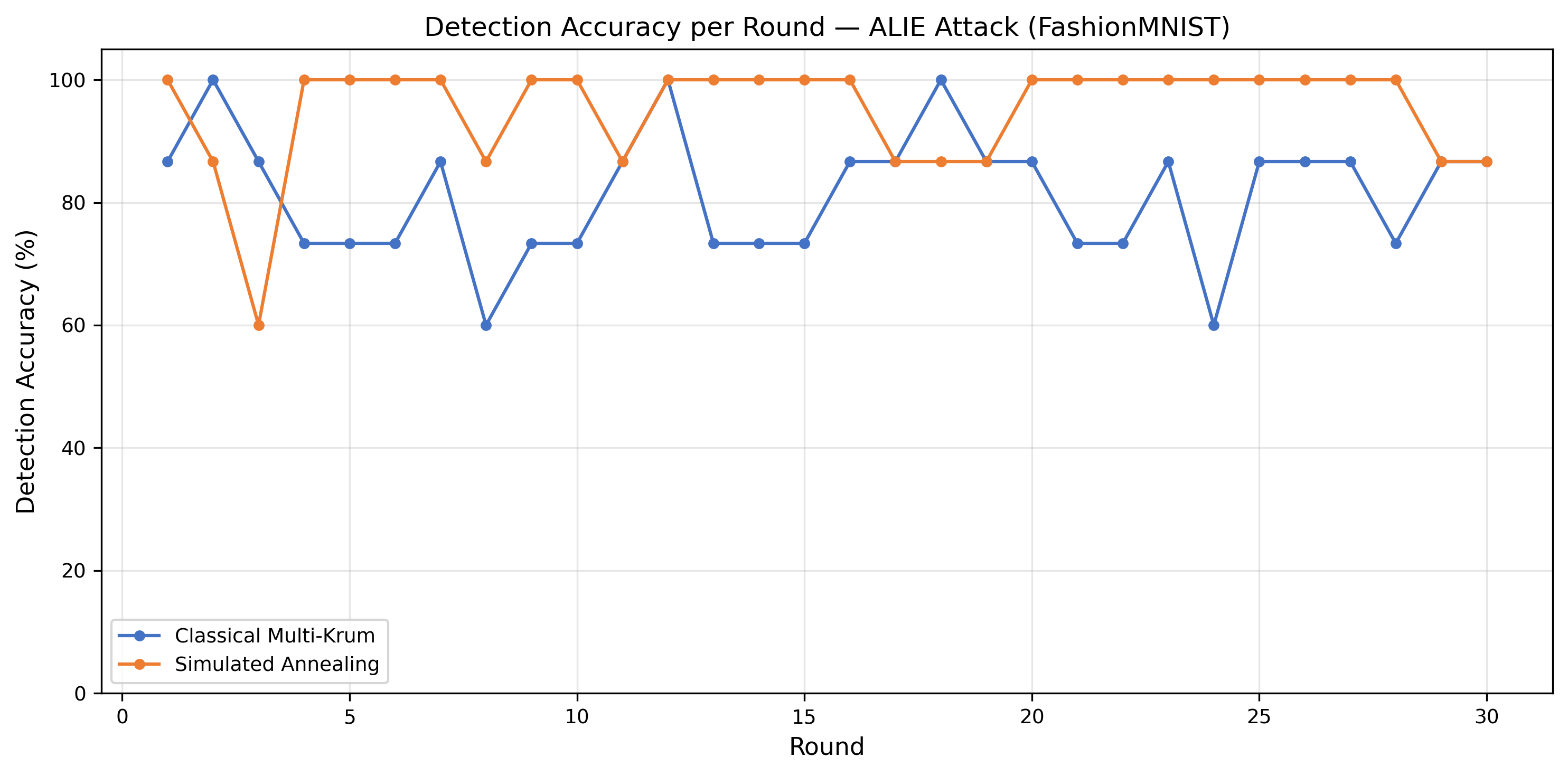}}
\caption{FashionMNIST: Round-by-round detection under ALIE attack.}
\label{fig:fmnist_advanced_lie}
\end{figure}

\begin{figure}[t]
\centerline{\includegraphics[width=\linewidth]{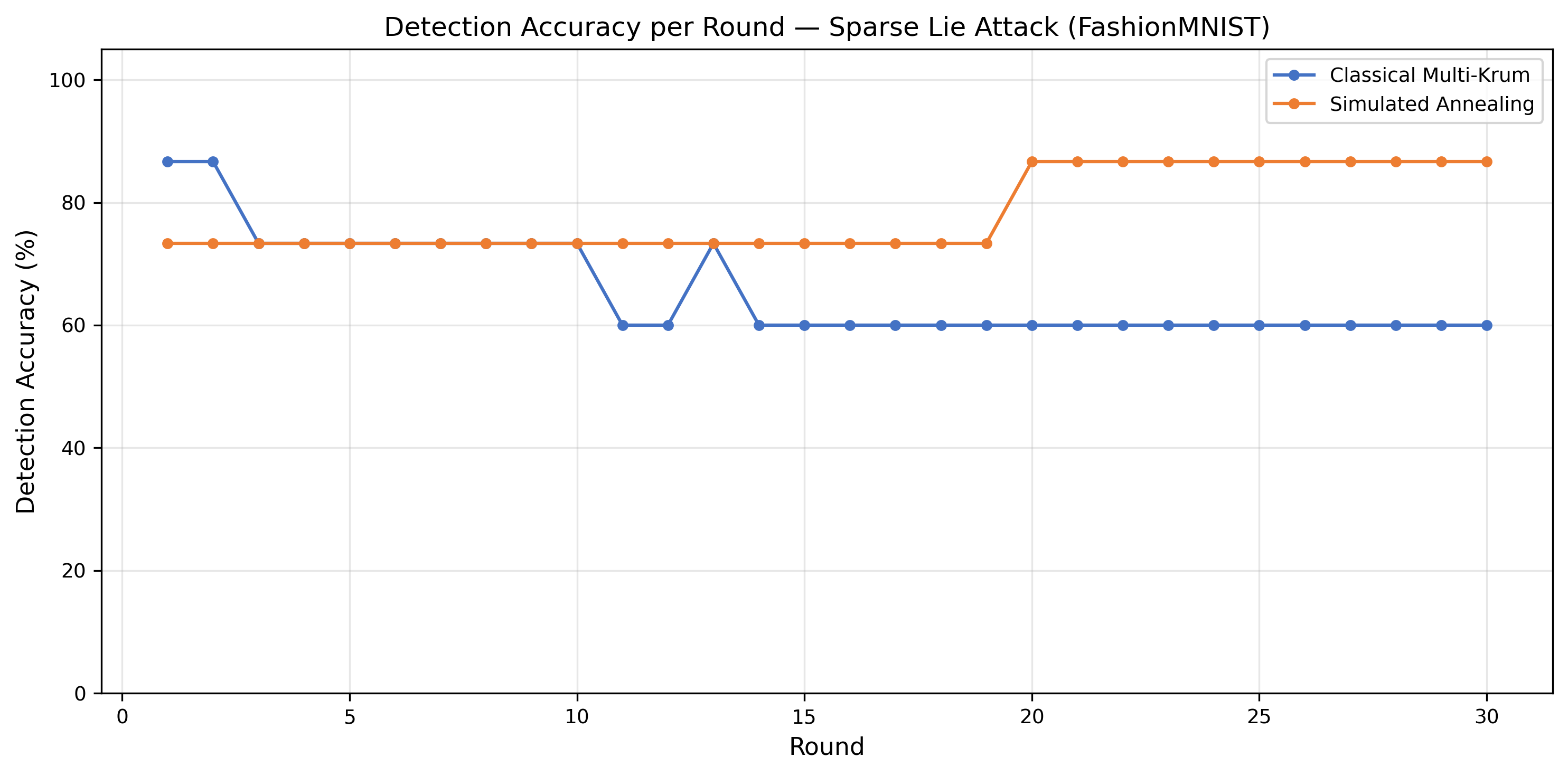}}
\caption{FashionMNIST: Round-by-round detection under Sparse Lie attack.}
\label{fig:fmnist_sparse_lie}
\end{figure}

Figures~\ref{fig:advanced_lie},~\ref{fig:sparse_lie},~\ref{fig:cifar_advanced_lie}, ~\ref{fig:cifar_sparse_lie},~\ref{fig:fmnist_advanced_lie}, and~\ref{fig:fmnist_sparse_lie} present round-by-round detection performance on MNIST, CIFAR-10, and FashionMNIST on ALIE and Sparse Lie attacks, clearly showing the benefits of QUBO formalism versus classical MultiKrum. Notably, the very similar curvature of the MNIST and FashionMNIST diagrams suggests that the detection is architecture rather than data driven.

\subsection{100-Client Hybrid Annealer Results}
Table~\ref{tab:100client_results} presents average detection accuracy over 5~rounds at 100~clients on MNIST, comparing classical MultiKrum, the initial Cascaded Dual-QUBO ensemble, and the MultiSignal ensemble. F1~scores are shown in Table~\ref{tab:100client_f1}.

\begin{table}[t]
\caption{MNIST: 100-Client Average Detection Accuracy averaged over 5~rounds with 80~honest, 20~Byzantine clients. Bold indicates the best method; ties are all bold.}
\label{tab:100client_results}
\centering
\small
\begin{tabular}{l|ccc}
\toprule
\textbf{Attack} & Classical & Cascade & MultiSignal \\
\midrule
Clustered            & \textbf{100.0} & \textbf{100.0} & \textbf{100.0} \\
Sign Flip            & \textbf{100.0} & \textbf{100.0} & \textbf{100.0} \\
ALIE         & 80.4           & 80.0           & \textbf{85.2}  \\
Sparse Lie           & 72.0           & 80.0           & \textbf{95.2}  \\
Gaussian Noise       & \textbf{100.0} & \textbf{100.0} & \textbf{100.0} \\
Scale                & \textbf{100.0} & \textbf{100.0} & \textbf{100.0} \\
\midrule
\textbf{Average}     & 92.07           & 93.33           & \textbf{96.73}  \\
\bottomrule
\end{tabular}
\end{table}

\begin{table}[t]
\caption{MNIST: 100-Client Average F1-score averaged over 5~rounds with 80~honest, 20~Byzantine clients. Bold indicates the best method; ties are all bold.}
\label{tab:100client_f1}
\centering
\small
\begin{tabular}{l|ccc}
\toprule
\textbf{Attack} & Classical & Cascade & MultiSignal \\
\midrule
Clustered            & \textbf{100.0} & \textbf{100.0} & \textbf{100.0} \\
Sign Flip            & \textbf{100.0} & \textbf{100.0} & \textbf{100.0} \\
ALIE         & 51.0  & 50.0         & \textbf{63.0}  \\
Sparse Lie           & 30.0   & 50.0        & \textbf{88.0}  \\
Gaussian Noise       & \textbf{100.0}& \textbf{100.0} & \textbf{100.0} \\
Scale                & \textbf{100.0}& \textbf{100.0} & \textbf{100.0} \\
\midrule
\textbf{Average}     & 80.17  & 83.33        & \textbf{91.83}  \\
\bottomrule
\end{tabular}
\end{table}

The MultiSignal ensemble achieves its most dramatic gain on Sparse Lie with 95.2\% detection accuracy vs.\ 72.0\% in classical MultiKrum, a 23.2 percentage-point gain, and vs.\ 80.0\% cascade, a 15.2 percentage-point gain. The suspicion penalty is particularly effective here because Sparse Lie modifies only the top 5\% highest-variance coordinates, producing Byzantine gradients that are near-identical to honest ones in 95\% of dimensions, which is exactly the pattern the $\tau_s$ threshold (10th percentile of pairwise distances) is designed to flag.

ALIE detection accuracy improves from 80.4\%  in classical MultiKrum and 80.0\% in cascade to 85.2\%. The agreement voting mechanism is effective here. The suspicion QUBO detects different subsets of Byzantine clients than classical MultiKrum, and combining both signals via QUBO-primary voting yields a selection better than either method alone.

Overall, the MultiSignal ensemble achieves 96.73\% average detection accuracy versus 92.07\% for classical MultiKrum and 93.33\% for the cascade, with perfect detection on all outlier attacks. The average F1~score improves from 80.17\% using classical MultiKrum to 91.83\% using MultiSignal, with the largest F1 gain on Sparse Lie from 30.0\% to 88.0\%.

\section{Discussion}

\subsection{Why the Approaches Differ}

The classic and QUBO-based solutions offer complementary approaches for assessing the trustworthiness of client updates.

Classical MultiKrum evaluates each client independently by computing its Krum score and greedily selects the $m$ clients with the lowest individual scores. Any client whose gradients look normal will be accepted, and any outlier will be flagged as Byzantine. Classical MultiKrum will identify the attacks like Gaussian Noise, Scale, Blatant Lie, Same Value, Targeted, Clustered, Label Flip, and Stealthy without fail.

QUBO-based selection jointly optimizes over all $\binom{n}{m}$ possible subsets using cosine distance. This makes it sensitive to the direction the gradient points, not its magnitude. The quadratic interaction terms $Q_{ij} = -2D_{ij} + 2\lambda$ encode pairwise relationships globally, picking the tightest cluster of $m$ clients. As a result, the QUBO method catches updates that, although not outliers from a Krum score perspective, have their gradients pointing in the wrong direction. QUBO-based selection is particularly effective with Lie, Advanced Lie, Sparse Lie, and Sign Flip, as these produce Byzantine gradients that appear statistically similar to honest gradients when evaluated individually. As our experiments showed, model architecture also plays a major role in adversarial attack detection. For example, the CNN used for CIFAR-10 encodes learned spatial filter patterns and the cosine signature changes are easily detectable. As a result, QUBO performed significantly better under the Shuffle attack (90.67\% vs.\ 70.22\%). Conversely, the same attack is not identified as well by QUBO when the model is significantly simpler, e.g., MNIST and FashionMNIST at $\sim$17K parameters, as these do not create distinctive enough directional anomalies. ALIE attack detection was particularly effective using QUBO. This attack modifies only 20\% of coordinates with large deviations, leaving 80\% unchanged; although it does not produce a significant Euclidean outlier, it shifts the gradient direction sufficiently to be detected by the QUBO formalism.
It is a well-known problem that annealing will perform sub-optimally given a complex energy landscape. As a result, given scattered Byzantine gradients, the QUBO objective's global minimum itself likely includes Byzantine updates.

\subsection{Complexity Analysis}

Classical MultiKrum runs in $O(n^2 d + n \log n)$, i.e., quadratic in clients $n$ times gradient dimension $d$ for pairwise distances, plus $O(n \log n)$ for sorting.

The QUBO pipeline has complexity $O(n^2 k + \gamma)$, where $k$ is the projection dimension and $\gamma$ the annealer time. For simulated annealing, $\gamma$ depends on the number of reads. For quantum  annealer, it includes embedding and QPU time, bounded by hardware rather than input size. With $n = 15$, the QUBO has 15 binary variables and $\binom{15}{2} = 105$ quadratic terms, well within current hardware capabilities.

Both Cascaded Dual-QUBO and MultiSignal ensemble computes both Euclidean Krum on full gradients, $O(n^2 d)$, and cosine Krum scores on projected gradients, $O(n^2 k)$, costing $O(n^2 d + n^2 k)$. For Cascaded Dual-QUBO, classical MultiKrum was used above gap threshold at a cost of $O(n^2 d)$. For attacks with below threshold gap, QUBO costs $O(n^2 k + \gamma)$.

Similarly for MultiSignal ensemble, classical MultiKrum is applied to outliers at a cost of $O(n^2 d)$. For the evasion route the dual distance matrix costs $O(n^2 k)$, QUBO construction and solving costs  $O(n^2 k + \gamma)$. Cascaded Dual-QUBO and MultiSignal ensemble thus incur no additional cost over the pure QUBO pipeline.


\subsection{Limitations}
We evaluated only three datasets (MNIST, FashionMNIST, CIFAR-10) and all models used in these experiments are relatively small. Larger models may require larger projection dimensions, which may not be feasible given current QA hardware limitations.
The experiments were performed with each of the 13 individual attack types in isolation, and the effects of a mixture of attacks have yet to be evaluated. In such a scenario, the routing gate could become confused. For example, if 50\% of attacks are Gaussian and another 50\% are ALIE, the wrong regime could be selected. A potential solution in such scenarios would be to switch to a ``paranoid'' mode, in which the union of all clients selected by both methods would be flagged as Byzantine.

\section{Conclusion}
We present in this paper a novel QUBO-based method for Byzantine-robust aggregation in FL. We evaluated our approach at two scales: first, a naive QUBO formulation using simulated annealing with 15 clients across three datasets; second a Cascaded Dual-QUBO ensemble that used a simple threshold-based routing mechanism to process outlier attacks using classical MultiKrum and evasion attacks using QA. 
As the Cascaded Dual-QUBO performance did not improve on ALIE, we experimented with a novel MultiSignal ensemble approach. 

We showed that for 15 clients, the QUBO formulation achieves up to 97.78\% detection accuracy on ALIE on CIFAR-10 and 95.11\% on MNIST, with an average 8 to 10 percentage-point improvement on the five hardest evasion attacks across all three datasets (Table~\ref{tab:evasion_summary}). The MultiSignal ensemble scales to 100~clients with 96.73\% average detection accuracy, a 4.7 percentage-point improvement, achieving Sparse Lie detection of 95.2\%, a 23.2 percentage-point improvement, and ALIE of 85.2\%, a 4.8 percentage-point improvement. The suspicion-penalized QUBO directly counteracts evasion attacks that exploit the standard QUBO's cluster-seeking objective, yielding the largest F1 improvement on Sparse Lie.

Future work will focus on scaling to over 500 clients by leveraging the QA's hybrid solver capacity. We aim to validate the existing method and enhance it as needed for mixtures of attacks. We also plan to evaluate wall-clock quantum advantage through direct QPU comparison, combining QUBO-based selection with quantum embedding of features~\cite{Ferenczi2025QE} in a unified quantum pipeline. We wish to extend our study to learn routing functions that adapt thresholds to the observed gradient distribution and investigate adaptive attacks specifically targeting the suspicion-penalized QUBO formulation.

\bibliographystyle{IEEEtran}
\bibliography{references}

\end{document}